\documentclass[10pt,twocolumn,letterpaper]{article}

\usepackage{wacv}             

\usepackage{graphicx}
\usepackage{amsmath}
\usepackage{amssymb}
\usepackage{booktabs}
\usepackage{enumitem}
\usepackage[pagebackref,breaklinks,colorlinks]{hyperref}
\usepackage{xcolor}
\usepackage[capitalize]{cleveref}

\crefname{section}{Sec.}{Secs.}
\Crefname{section}{Section}{Sections}
\Crefname{table}{Table}{Tables}
\crefname{table}{Tab.}{Tabs.}

\def\wacvPaperID{687}
\def\confName{WACV}
\def\confYear{2025}

\begin{document}

\title{EgoPoints: Advancing Point Tracking for Egocentric Videos}

\author{Ahmad Darkhalil$^{1}$ \quad Rhodri Guerrier$^{1}$ \quad Adam W. Harley$^{2}$ \quad Dima Damen$^{1}$ \\
$^{1}$University of Bristol \quad
$^{2}$Stanford University\\
\small{\url{http://ahmaddarkhalil.github.io/EgoPoints}}
}

\maketitle

\begin{figure*}[!h]
  \centering
    \includegraphics[width=1.0\linewidth]{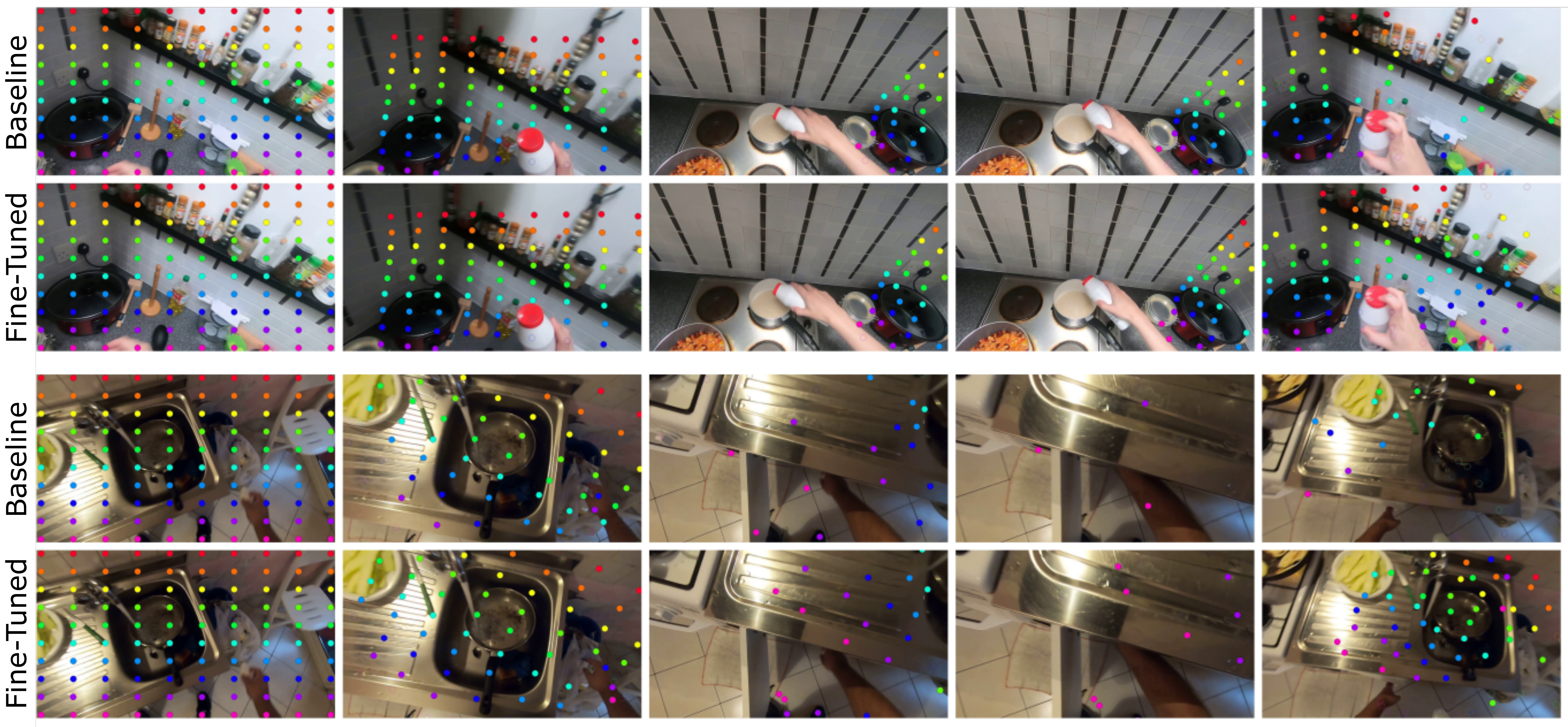}
    \vspace*{-24pt}
    \caption{Sample sequences from EgoPoints, with dense points in reference frame (left) tracked through head motion where both scene points and dynamic object points leave the field of view and and return during the sequence -- e.g. in the first row, the salt bottle is used in a different part of the scene then returned back to shelf. We show qualitative results of CoTracker \cite{karaev2023cotracker}, before and after fine-tuning with our synthetic sequences combining Kubric and EPIC Fields points (K-EPIC). Fine-tuning increases the number of re-identified.}
    \vspace*{-12pt}
  \label{fig:main_qual}
\end{figure*}

\begin{abstract}
We introduce EgoPoints, a benchmark for point tracking in egocentric videos.
We annotate 4.7K challenging tracks in egocentric sequences. Compared to the popular TAP-Vid-DAVIS~\cite{doersch2022tap} evaluation benchmark, we include 9x more points that go out-of-view and 59x more points that require re-identification (ReID) after returning to view.
To measure the performance of models on these challenging points, we introduce evaluation metrics that specifically monitor tracking performance on points in-view, out-of-view, and points that require re-identification.

We then propose a pipeline to create semi-real sequences, with automatic ground truth. We generate 11K such sequences by combining dynamic Kubric~\cite{greff2022kubric} objects with scene points from EPIC Fields \cite{tschernezki2024epic}.
When fine-tuning point tracking methods on these sequences and evaluating on our annotated EgoPoints sequences, we improve CoTracker \cite{karaev2023cotracker} across all metrics, including the tracking accuracy $\delta^\star_{\text{avg}}$  by 2.7 percentage points and accuracy on ReID sequences (ReID$\delta_{\text{avg}}$) by 2.4 points. We also improve $\delta^\star_{\text{avg}}$ and ReID$\delta_{\text{avg}}$ of PIPs++\cite{zheng2023pointodyssey} by 0.3 and 2.8 respectively.
\vspace*{-12pt}
\end{abstract}

\section{Introduction}
\label{sec:intro}
Given a sequence of frames and a set of query point coordinates in the first frame, the task of point tracking is to output the coordinates of each query point in all of the subsequent frames. Point tracking in general has seen tremendous success thanks to the introduction of large-scale synthetic training datasets~\cite{greff2022kubric, zheng2023pointodyssey}, custom algorithms~\cite{harley2022particle, zheng2023pointodyssey, karaev2023cotracker} and carefully annotated videos from YouTube~\cite{doersch2022tap}. Tracking these points can be particularly challenging when they are occluded or out-of-view. Therefore, previous works \cite{karaev2023cotracker,karaev2024cotracker3, doersch2022tap, doersch2023tapir, cho2024local} learn to track through short occlusions by using multi-frame models and predicting point visibility.

Standard point tracking datasets~\cite{doersch2022tap} typically annotate subjects captured from a third person perspective where the camera motion is either limited or smooth. Objects remain mostly in-view and rarely leave the field of view during these sequences.
It is therefore unclear whether these methods can also work on egocentric videos---videos captured from head-mounted cameras while the camera wearer is performing actions.
Standard tasks such as object tracking~\cite{NEURIPS2023_ef01d91a,dunnhofer2023visual}, video object segmentation~\cite{darkhalil2022epic} and camera pose estimation~\cite{EPICFields2023} have shown to be more challenging when tested on egocentric videos~\cite{damen2022rescaling,grauman2022ego4d}. 
This is due to the fast camera motion, objects leaving and returning to the field of view regularly and high levels of occlusion. 

For the first time, we address the task of dense point tracking in egocentric videos.
Point tracking in egocentric videos has significant potential for human-to-robot imitation learning~\cite{Vecerik24Short}, as well as projecting both foreground and background assets in augmented reality.
We first annotate just over 500 egocentric video clips from EPIC-KITCHENS-100~\cite{damen2022rescaling} for sparse point tracking to evaluate current methods.
We demonstrate that performance on these sequences is significantly lower than in existing public point tracking benchmarks (Fig~\ref{fig:main_qual}).
We analyse reasons for failure, showcasing the frequent cases where objects leave view and return, as both head motion and actions cause these effects. 
We introduce metrics that are necessary to quantify the performance of these methods on egocentric sequences.

Finally, inspired by prior works \cite{zheng2023pointodyssey, karaev2023cotracker}, we synthesise sequences that can be used to fine-tune these methods for increased performance.
We combine scene points extracted from the EPIC Fields dataset~\cite{EPICFields2023} with points on dynamic 3D objects from Kubric~\cite{greff2022kubric}.
Importantly, we prioritize the sampling of sequences that focus on re-identification of both static scene points and dynamic object points, using heuristics on camera trajectories and object motion, and also reversing sequences. Together, we refer to these synthetic sequences as K-EPIC, as each sequence is sampled from both Kubric and EPIC-KITCHENS.
We fine-tune two point tracking methods on these sequences, consistently improving performance on standard metrics as well as on our re-identification metrics.

To summarise, our contributions are as follows. (1) We identify challenges that state-of-the-art point trackers face in egocentric videos. (2) We propose a new benchmark and new metrics to showcase these challenges: we manually annotate 517 challenging egocentric sequences particularly targeting head motion and re-identification of dense points on both scene and dynamic objects. We refer to this benchmark as \textit{EgoPoints}. 
(3) To help make progress on these challenges, we propose a pipeline to generate synthetic training data through combining Kubric objects with sequences from EPIC Fields. 
After fine-tuning two models (PIPs++ and CoTracker) on this data, we show improved performance on EgoPoints, while maintaining performance on standard point tracking benchmarks. 

\section{Related Work}
\label{sec:related_work}

\noindent \textbf{Point tracking} is a classic task in computer vision \cite{lucas1981iterative,tomasi1991detection}. Much work in this area revolves around ``optical flow'', where the goal is to estimate the displacement field that tracks points from one frame to another. Early optical flow methods used optimization techniques\cite{Brox2011LargeDO,brox_densepoint}; more recent methods learned feed-forward models, supervised from synthetic datasets \cite{flownet,flownet2,raft}. Meanwhile, Sand and Teller~\cite{particlevideo} formalized the goal of tracking points across multiple frames, which adds the challenge of managing occlusions and achieving temporal coherence. 
Harley et al.~\cite{harley2022particle} recently revitalized this setup with PIPs, a multi-frame model for tracking points through occlusions, with an architecture inspired by state-of-the-art optical flow models. Many recent methods have built further in this direction, adding wider spatial awareness~\cite{bian2024context}, wider temporal awareness~\cite{zheng2023pointodyssey}, joint multi-point tracking~\cite{karaev2023cotracker}, 
patch-wise correlations~\cite{cho2024local} and optimization-based techniques~\cite{wang2023tracking,tumanyan2024dino}. These developments have been supported by concurrent work on producing training and testing data for the models. Synthetic training data has progressed from unrealistic to more realistic, beginning with random flying geometry~\cite{flyingthings16,harley2022particle} to physics-based rendering of random objects~\cite{greff2022kubric}, and most recently using motion capture as a driver for animated character motion~\cite{zheng2023pointodyssey}. Real-world datasets have so far been sampled from YouTube~\cite{doersch2022tap} and from driving data~\cite{balasingam2024drivetrack}. Recent works have utilised boot-strapping techniques on unlabelled videos~\cite{sun2024refining, doersch2024short,karaev2024cotracker3} to improve the performance of pre-trained point trackers. However, existing benchmarks lack egocentric data. A concurrent work~\cite{koppula2024tapvid} focused on 3D point tracking and acknowledged this deficiency, introducing a benchmark from diverse videos including egocentric ones. However, this was collected from model kitchens~\cite{Pan_2023_ICCV}, whilst our data comes from natural environments. 

\noindent \textbf{Egocentric vision} focuses on understanding the actions and the interactions of a camera wearer, by capturing the activities from their first view perspective.
Traditionally tested in controlled settings and small-scaled datasets~\cite{fathi2011learning,5204354}, the field was fuelled by large-scale data collections~\cite{damen2022rescaling,grauman2022ego4d,Grauman_2024_CVPR} with benchmarks for long-term object tracking~\cite{huang_etal_cvpr23,NEURIPS2023_ef01d91a,dunnhofer2023visual}, pixel-level object segmentation~\cite{darkhalil2022epic,tokmakov2023breaking}, body and hand pose~\cite{Khirodkar_2023_ICCV,Grauman_2024_CVPR} amongst other dense tasks.
All these works highlight the challenges related to significant camera motion and objects frequently leaving the field of view and appearing again. In the EgoTracks dataset~\cite{NEURIPS2023_ef01d91a}, where bounding boxes of objects were annotated from egocentric videos, the authors note significant challenges posed by frequent disappearance and re-appearance of objects, similar to our work, but our annotations are at the point level.  

\section{EgoPoints: Evaluation Benchmark}
\label{sec:manual_dataset}

In this work, we offer the first attempt to assess point tracking on unscripted egocentric videos.
In this section we introduce our EgoPoints benchmark, which we believe covers specific important weaknesses of prior datasets. 

To contextualize our benchmark, we first review existing datasets. The current datasets used in point tracking fall under two main categories: synthetic and real. Synthetic datasets are often used for training but also can be part of test sets, such as PointOdyssey~\cite{zheng2023pointodyssey} and TAP-Vid-Kubric~\cite{doersch2022tap}. Datasets with real world video on the other hand are often used exclusively for testing, due to their small size and the difficulty of annotating points.
For example, TAP-Vid-DAVIS~\cite{doersch2022tap} contains 30 annotated videos from the DAVIS 2017 validation set \cite{pont20172017}, and has a small number of annotations per video.
TAP-Vid-KINETICS~\cite{doersch2022tap} contains more lighting and motion diversity across a larger set of 1189 videos, but it only has tracks of at most 250 frames, which is relatively short-term. In both, the camera motion remains limited or smooth, with points typically remaining within the image bounds (even if occluded). Additionally, sequences tend to be short-term. 

We believe that the large variety of available egocentric data could be perfect to fill the gaps in existing point tracking data, as these videos are often filmed in messy indoor environments, over long time periods and contain a lot of movement of the head, hands and the objects. We therefore propose a new manually annotated point benchmark of videos taken from the popular EPIC-KITCHENS-100 dataset \cite{damen2022rescaling} with a particular emphasis on including sequences requiring point re-identification (i.e., where a point leaves view and then returns). This we believe is an important and overlooked aspect of point tracking methods today and is discussed at length in Section~\ref{sec:point_track_failures}. Due to the costly aspect of point annotation, we propose a sparsely annotated set. This allows us to focus on the areas of re-identification whilst also maintaining tracking accuracy and precision.

We start our data collection process by using the VISOR annotations of EPIC-KITCHENS\cite{darkhalil2022epic}. These annotations allow us to automatically identify sequences where the camera wearer's hands leave and re-enter the field of view of the camera.
These are good indicators of dynamic objects being moved around.
Secondly, we use point clouds from EPIC Fields \cite{tschernezki2024epic} to automatically determine sequences where the camera is observing part of the 3D scene, moving to another part then re-observing the original part. We use common in-view points, from the sparse point clouds, to find these sequences. 
By combining both, we identify sequences where point tracking requires re-identifying scene points that are part of the environment, as well as dynamic objects that re-appear at different times/locations.
These offer the most challenging cases for dense point tracking as algorithms need to handle neighbouring points that are moving independently.
We select two settings -- 
one that maximises re-identification where the head motion moves to another part of the scene and back (250 sequences) and another where the head motion keeps some of the points in the field of view at all times (267 sequences).

\begin{table*}[t]
\centering
\resizebox{0.7\linewidth}{!}{
\begin{tabular}{lccccc}
\toprule
Dataset       & Total Tracks & OOV Tracks & ReID Tracks & Avg. Video Length & Avg. Points/Frame \\
\midrule
TAP-Vid-DAVIS & 650  & 94  & 10 & 66.6 & \textbf{21.7}  \\
EgoPoints      & \textbf{4703}  & \textbf{875} & \textbf{593} & \textbf{511.0} & 8.5 \\
\bottomrule
\end{tabular}}
\vspace*{-2pt}
\caption{Comparisons of our annotated sequences, EgoPoints, and the commonly used TAP-Vid-DAVIS \cite{doersch2022tap} point tracking benchmarks}
\vspace*{-12pt}
\label{tab:manual_data_comp}
\end{table*}

\begin{figure}[t]
  \centering
   \includegraphics[width=\linewidth]{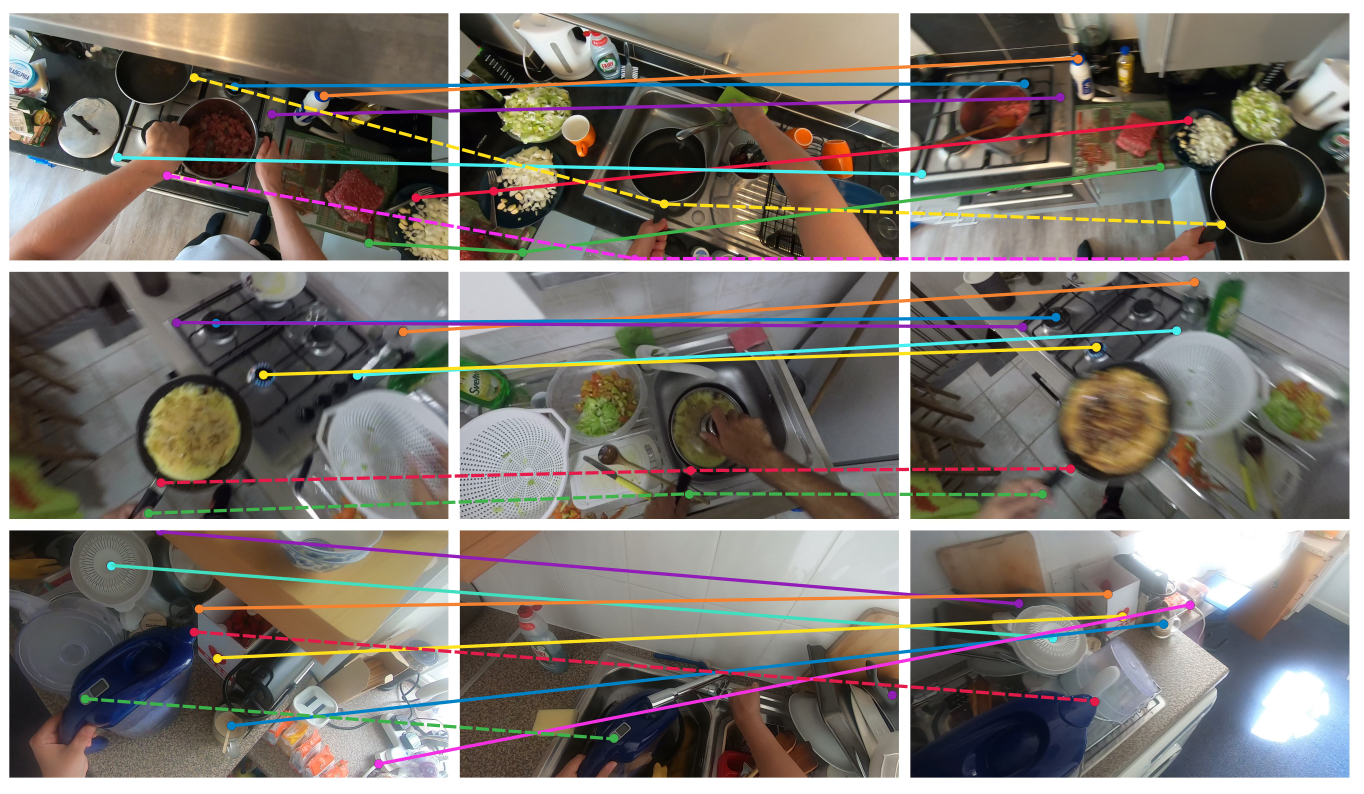}
   \caption{Example of sparsely annotated sequences from EgoPoints benchmark (annotated at 1080p full res images). We expand the point/pixel radius expanded for purposes of visualisation. The dashed lines represent dynamic object tracks, while solid lines show scene point tracks.}
   \vspace*{-12pt}
   \label{fig:point_annotator}
\end{figure}

For each sequence, we manually select three images from the sequence (one reference and two evaluation frames). 
These are selected to cover the extent of the sequence but also capture both dynamic and static re-identified objects.
We work with a single expert annotator for consistency and use a custom-built interface as follows.
All images in the interface are of 1920x1080 resolution and a zoom functionality is provided to enable accurate pixel selection.
The annotator is instructed to select roughly 10 points visible in the reference frame which are also visible in at least one of the two evaluation frames. For each evaluation frame, the annotator is instructed to either annotate the corresponding point location or flag whether the point is out-of-view, or in-view but occluded.
We use these flags to evaluate the ability of methods to track points out-of-view as well as re-identify points.
There is also functionality for the annotator to label whether a point is part of the static scene or on a dynamic object.

Figure \ref{fig:point_annotator} showcases three sample sequences from EgoPoints, overlaid with annotated points, which are coloured and connected for visualization. These showcase the challenging re-identification scenarios of both scene and object points. For example, in the first row, the track of the annotated point on the pan handle (dotted yellow) shows the object moved across the kitchen with the pan rotated and tilted. The scene points (e.g. on the hob) leave the view in the middle frame and re-appear in the last frame.

\begin{figure}[t]
  \centering
   \includegraphics[width=\linewidth]{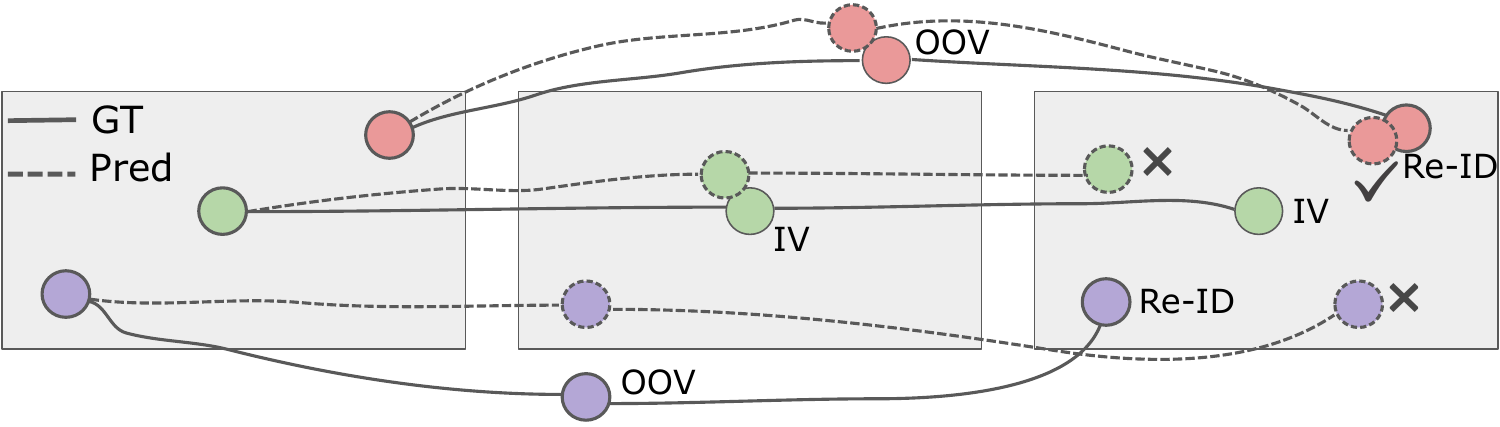}
   \vspace*{-20pt}
   \caption{Visualisation of three points tracked over three frames classified by the metrics in EgoPoints. IV: in-view, OOV: out-of-view, ReID: Re-identification (in-view after being out-of-view). $\checkmark$: correctly tracked, $\text{x}$: incorrectly tracked.}
   \vspace*{-16pt}
   \label{fig:re_id_metrics}
\end{figure}

In total, our benchmark contains 517 sequences with 4703 tracks. The average length of a video is 511.0 frames and the average point count per video is 8.5. Of these tracks, EgoPoints includes 875 tracks with an out-of-view point and 593 re-identification scenarios (or ReIDs). These categories are best understood with aid from the visualisation in Figure~\ref{fig:re_id_metrics}. A point from the reference frame is manually flagged to be out-of-view at an evaluation frame when the annotator notes the object or part of the scene has left the field of view between the two frames. This is distinct from occluded or invisible points, which could still be within the field of view of the camera but invisible. We also define a ReID flag for when the point is out-of-view in an intermediate frame, then in-view at a later frame. These are particularly challenging tracks that we wish to highlight in our benchmark. The red and purple ground truth tracks of Figure~\ref{fig:re_id_metrics} are ReID cases. Of these, the red prediction is visualised as a correct ReID while the purple prediction is an incorrect ReID because the point in the second frame incorrectly remains in-view and the prediction is much further from the ground truth in the final frame.

A comparison of our EgoPoints dataset and the popular TAP-Vid-DAVIS dataset \cite{doersch2022tap} can be found in Table~\ref{tab:manual_data_comp}. As the DAVIS points do not have flags for whether invisible points are occluded (but in-view) or are out-of-view, we manually annotate these points for all 30 videos to allow this analysis. 

The dataset has many points on both scene and dynamic objects that go out-of-view and re-appear due to the nature of these videos. 
EgoPoints offers a much broader set of challenging tracks compared to TAP-Vid-DAVIS. 
Approximately 19\% of all tracks in EgoPoints contain a point that goes out-of-view in one of the two evaluating frames, and 13\% of tracks require re-identification of a point after it had left the scene (in contrast with 1.5\% of tracks in DAVIS).
Additionally, EgoPoints has a clear additional challenge in video length over DAVIS, with nearly eight times the average length. This is also important for assessing models as previous works have focused on short videos.

\section{Challenges for Current Models}
\label{sec:point_track_failures}

As discussed in the previous section, most existing point tracking evaluation videos are short-term recordings with relatively simple camera motion and very few targets leaving view or requiring re-identification.
To quantify the weaknesses of existing models, we evaluate five of the state-of-the-art models for point tracking by focusing on point re-identification. This is achieved using the manually annotated EgoPoints benchmark described in Sec~\ref{sec:manual_dataset}.

\begin{table}
\centering
\resizebox{\linewidth}{!}{
\begin{tabular}{lcrrrr}
\toprule
&TAP-Vid-DAVIS &\multicolumn{4}{c}{EgoPoints}\\
\cmidrule(lr){2-2} \cmidrule(lr){3-6}
Model & $\delta_{\text{avg}}\uparrow$ & $\delta_{\text{avg}}\uparrow$ &ReID$\delta_{\text{avg}}\uparrow$ & OOVA$\uparrow$ & IVA$\uparrow$ \\
\midrule
PIPs++ \cite{zheng2023pointodyssey} &64.0 & 36.9 & 14.6 & 50.4 & 89.2 \\
CoTracker \cite{karaev2023cotracker} & 74.7 & 38.5 & 4.8 & \textbf{81.4} & 73.4 \\
BootsTAPIR Online \cite{doersch2024bootstap} & 65.2 & 39.6 & 0.0 & 0.0 & \textbf{100.0} \\
LocoTrack \cite{cho2024local} & 75.3 & \textbf{59.4} & 0.1 & 0.2 & 99.9 \\
CoTracker v3 \cite{karaev2024cotracker3} & \textbf{77.2} & 50.0 & \textbf{15.0} & 31.8 & 99.3 \\
\bottomrule
\end{tabular}}
\caption{Performance of point tracking baselines on the commonly used benchmark TAP-Vid-DAVIS compared to our EgoPoints evaluation benchmark on main metric $\delta_{\text{avg}}$. Additionally, metrics showcasing ReID, out-of-view (OOVA) and in-view (IVA) accuracy are added to showcase where models fail. We use the query-first mode in all evaluations. }
\vspace*{-12pt}
\label{tab:sota_failures}
\end{table}

Table \ref{tab:sota_failures} highlights the performance of a handful of SOTA models, namely PIPs++ \cite{zheng2023pointodyssey}, CoTracker \cite{karaev2023cotracker}, BootsTAPIR~\cite{doersch2024bootstap}, LocoTrack~\cite{cho2024local} and CoTracker v3~\cite{karaev2024cotracker3}, 
on metrics that focus on tracking, occlusion and ReID performance. We briefly describe the used metrics here but defer the detailed explanation of each metric to Sec~\ref{sec:results}.
We first compare the performance on TAP-Vid-DAVIS to EgoPoints on the standard metric $\delta_{\text{avg}}$, which measures points correctly tracked within a predefined set of thresholds. 
The drop in performance when comparing TAP-Vid-DAVIS to EgoPoints is evident in every case: from 64.0 to 36.9 for PIPs++, from 74.7 to 38.5 for CoTracker, from 65.2 to 39.6 for BootsTAPIR Online, 75.3 to 59.4 for LocoTrack and 77.2 to 50.0 on the recently released CoTracker v3.

When measuring the ReID capability of all methods, results demonstrate poor performance, with almost all ReID points failing to be correctly tracked (only 14.2\% of points using PIPs++, 2.8\% with CoTracker, 15.0\% for CoTracker v3 and 0.0\% for both BootsTAPIR Online and LocoTrack).

PIPs++~\cite{zheng2023pointodyssey} tends to allow points to remain on the screen even when their ground truth tracks have left.
 It is much better at keeping points in the field of view (89.2\% in-view accuracy, or IVA), but this is at the cost of failing to track correctly as the point leaves the field of view (50.4\% out-of-view accuracy, or OOVA).
The recently released CoTracker v3 \cite{karaev2024cotracker3} has a similar behaviour.
This is in contrast to the original CoTracker \cite{karaev2023cotracker} which has a much better OOVA: points leave the field of view correctly, but then the model struggles to bring them back into frame, as identified by the lower IVA and poor ReID.
Other recent models (BootsTAPIR~\cite{doersch2024bootstap} and LocoTrack~\cite{cho2024local}) forcibly keep all point estimates in-view, resulting in near 0 performance in predicting points out-of-view (OOVA) and identify re-appearing points.
Their improved performance is a result of their focus on in-view points, which form almost all annotated points in previous benchmarks.

\begin{figure*}
  \centering
  \begin{subfigure}{0.45\linewidth}
    \includegraphics[width=1\linewidth]{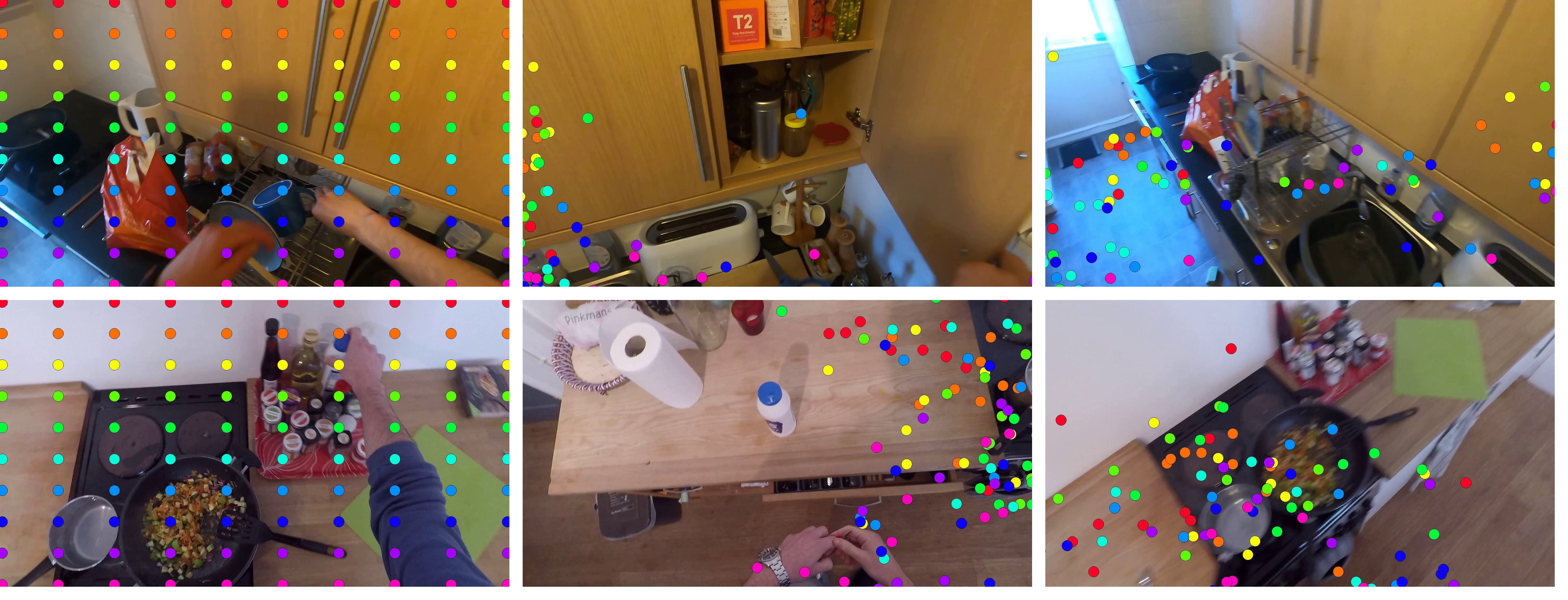}
    \caption{PIPs++ \cite{zheng2023pointodyssey}}
    \label{fig:pips_failures}
  \end{subfigure}
  \hfill
  \begin{subfigure}{0.45\linewidth}
    \includegraphics[width=1\linewidth]{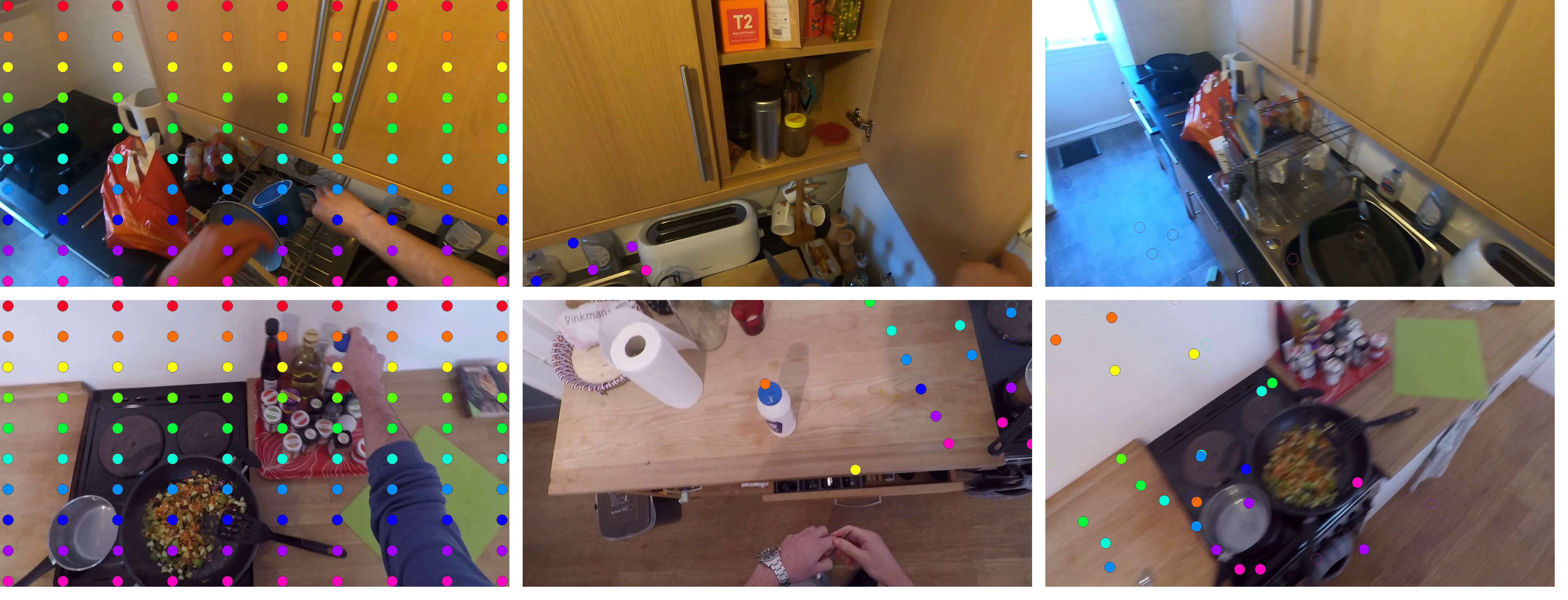}
    \caption{CoTracker \cite{karaev2023cotracker}}
    \label{fig:cotracker_failures}
  \end{subfigure}
  \vspace*{-8pt}
  \caption{Examples of re-identification failures in state-of-the-art models. Each row represents a particular video. The top sequence is 305 frames long, whilst the bottom sequence is 994 frames long.}
  \label{fig:re_id_failures}
\end{figure*}

Figure~\ref{fig:re_id_failures} showcases qualitative examples on the EgoPoints evaluation benchmark. 
On each row, we show a sparse grid of points on the first frame, which is the reference frame. 
As the camera moves to another scene (middle) then returns to the same scene (right), we show two methods' performance.
Both PIPs++ \cite{zheng2023pointodyssey} and CoTracker \cite{karaev2023cotracker} struggle to track the points out-of-view and re-locating them as they return. PIPs++ quickly loses the structure of the grid upon head turning and only manages to re-identify a few points because the points incorrectly float around on the screen during occlusion. In contrast, CoTracker retains the structure of the grid of points. This is likely due to the padding that the model uses, which allows it to track neighbourhoods of points as these leave the field of view. However, CoTracker fails to track these points back as the camera returns to the part of the scene from the reference frame.
We compare a short sequence on the first row (6s long) to a longer sequence (16s) on the second row.
The performance is evidently worse for the longer sequence.

The poor performance of current methods on the EgoPoint benchmark can be inherent in the algorithms: the overlapping sliding windows and feature matching approaches, or due to the training sets. We explore this by proposing a dataset for fine-tuning models so they are better adapted to challenges in egocentric videos.

\section{K-EPIC: Semi-Real Training Sequences}
\label{sec:k_epic_dataset}

In an attempt to address the poor performance of current models on the EgoPoints benchmark, we aim to fine-tune these models on data that addresses the challenges in this benchmark. We thus introduce a pipeline to produce automatically annotated data for training from available resources. 
These sequences are semi-real: the background (and scene points) are sampled from real egocentric video sequences, whilst the foreground objects come from synthetic 3D objects. We describe this pipeline next.

To produce the scene points, we utilise the point clouds and camera poses made available from EPIC Fields \cite{EPICFields2023}. 
We sample sequences of 32 frames with head motion around various parts of the scene. We sample these sequences by clustering the head motion (considering both translation and rotation) into 3 clusters and selecting the sequences in the cluster with the most significant head motion.

To ensure the reliability of scene points, we filter out noisy 3D points and those which project onto dynamic objects, as all scene points should be located on scene elements that remain static throughout the sequence. To achieve this we employ a pretrained CoTracker model \cite{karaev2023cotracker}.
We explain these steps in more detail next. 

\vspace{1em}

\noindent \textbf{A. Project and Track.}
After selecting sequences with sufficient head motion, we project 3D scene points from the EPIC Fields \cite{EPICFields2023} point clouds onto the first frame of the sequence. We only keep points that have a minimum track length of 20 frames, and a maximum re-projection error of 1 pixel (automatically calculated by COLMAP~\cite{schoenberger2016sfm} while reconstructing the scene), and limit to 4,000 points.

\begin{figure*}
  \centering
    \includegraphics[width=1.0\linewidth]{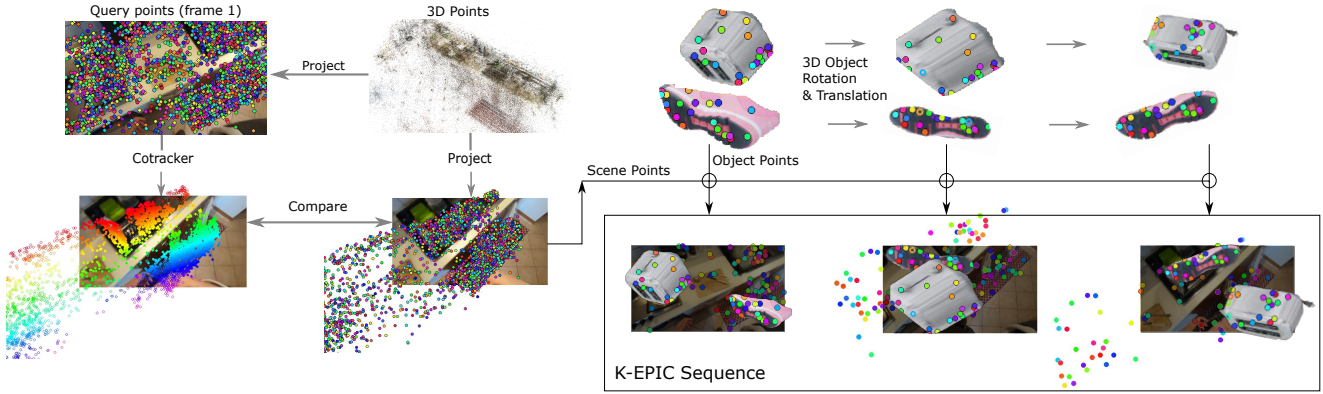}
    \vspace*{-12pt}
    \caption{The pipeline for K-EPIC. This includes projecting 3D points as tracks and filtering them using CoTracker to get scene points (left). Additionally, we sample 3D objects and tracks from TAP-Vid-KUBRIC (top right). These are combined to produce K-EPIC sequences with ground-truth point tracking. The number of sampled points and brightness of the images are decreased for visualisation purposes.}
    \vspace*{-12pt}
  \label{fig:kepic_pipeline}
\end{figure*}

\noindent \textbf{B. Compare and Filter.}
To further remove noise, we use the assumption that these points are static in 3D. We run CoTracker (using a checkpoint trained on TAP-Vid-KUBRIC~\cite{doersch2022tap}) on these sequences and filter away points whose projected motion is inconsistent with CoTracker.
Notice that camera estimates also inform us about whether the points are in or out-of-view in each frame. 
We thus only compare trajectories for in-view frames.
We use an L2 distance threshold of 1 pixel between projected motion and CoTracker, to remove noisy points or those that are on dynamic objects. Once this pruning of points is complete, we are left with confident `scene' points that can be used for training. Additionally, we use CoTracker's visibility estimates as pseudo ground truth when the points are in view.

\noindent \textbf{C. Add Dynamic Objects.}
Static scene points are not enough to help with the tracking and re-identification of dynamic scene points (as we later demonstrate in Table~\ref{tab:sota_improv}). To train for dynamic objects, we require ground truth 3D objects with pose changes of manipulated objects. We use synthetic 3D objects and their points from the popular TAP-Vid-KUBRIC \cite{doersch2022tap} dataset: a synthetic training dataset with objects falling and colliding with each other. 
We extract the objects from the already created TAP-Vid-KUBRIC sequences, where roughly 20 objects have been sampled from a library of 3D objects, rotated across a random axis and translated spatially.
Due to the restrictions of TAP-Vid-KUBRIC, these sequences are only 24 frames long. 

\noindent \textbf{D. Combine Scene and Object Points} We overlay the object points on top of the temporally-resampled scene points, producing new 24-frame training sequences. 

In total, we generate 11K sequences with an average of 2008 tracks per sequence. The total number of tracks is 22.1M, of which 10M are scene points and 12.M is sampled on the moving 3D foreground objects with an average of 9.2 objects per sequence. Our scene points have an average re-projection error of 0.45 pixels and a mean track length of 364 frames. Additionally, they have a median point speed of 8.0 pixels/frame compared to 2.9 for the scene points of TAP-Vid-KUBRIC~\cite{doersch2022tap} calculated after resizing images into 854x480 pixels. Figure \ref{fig:kepic_pipeline} summarize the overall pipeline to generate a sample sequence of K-EPIC.

We randomly augment the K-EPIC sequences with looped versions, where frames are re-ordered as: [1,3,5,..,21,23,24,22,20,...,4,2]. This returns the objects to their original locations, forcing re-identification. 

\section{Results}
\label{sec:results}

\begin{table*}
\centering
\resizebox{0.9\linewidth}{!}{
\begin{tabular}{lccrcccr}
\toprule
& \multicolumn{3}{c}{$\delta$ 
 Metrics} & \multicolumn{3}{c}{Accuracy Metrics} & Error \\
\cmidrule(lr){2-4} \cmidrule(lr){5-7} \cmidrule(lr){8-8}
Model      & $\delta_{\text{avg}}\uparrow$ & $\delta_{\text{avg}}^*\uparrow$ &  ReID$\delta_{\text{avg}}\uparrow$ & IVA$\uparrow$&OOVA$\uparrow$  & OA$\uparrow$ & MTE$\downarrow$       \\
\midrule
PIPs++ \cite{zheng2023pointodyssey}    & \textbf{36.9} & 57.8 & 14.0 & 89.2 & 50.4 & -- & 22.9 \\
PIPs++ w. K-EPIC FT (scene points only) & 36.3 & 57.8 & 13.0 & \textbf{90.1} & \textbf{53.0} & -- & 22.9 \\
PIPs++ w. K-EPIC FT (scene and object points) & 36.6 & \textbf{58.1} & \textbf{16.8} & 89.9 & 52.0 & -- & \textbf{22.2} \\

\midrule
CoTracker \cite{karaev2023cotracker} & 38.5 & 54.8 & 4.8 & 73.4 & 81.4 & 81.0 & 52.1 \\
CoTracker w. K-EPIC FT (scene points only) & 38.9 & 56.0 & 6.3 & 74.8 & \textbf{85.4} & 80.7 & 51.3 \\
CoTracker w. K-EPIC FT (scene and object points) & \textbf{39.6} & \textbf{57.5} & \textbf{7.2} & \textbf{78.1} & 82.0 & \textbf{81.8} & \textbf{40.5} \\
\bottomrule           
\end{tabular}}
\caption{Performance improvement on EgoPoints after fine-tuning on only scene points vs K-EPIC (scene and points). FT means the fine-tuned version of the model}
\vspace*{-12pt}
\label{tab:sota_improv}
\end{table*}

Due to time constraints, we do not include fine-tuning experiments for all models in Table \ref{tab:sota_failures}. Instead, we pick the two established models from our baseline results. Specifically, we use the publicly released checkpoints from two state-of-the-art point trackers for our experiments: PIPs++~\cite{zheng2023pointodyssey} trained on  PointOdyssey~\cite{zheng2023pointodyssey} and  CoTracker~\cite{karaev2023cotracker} trained on TAP-Vid-KUBRIC~\cite{doersch2022tap}.
We fine-tune these models on K-EPIC, using batches where two-thirds are from K-EPIC, and one-third is from the previous training dataset for the corresponding checkpoint.
This minimises the catastrophic forgetting when testing on non-egocentric datasets.
For PIPs++, we fine-tune using one V100 32GB GPU for 45K iterations. For CoTracker, we fine-tune using two V100s for 34K iterations.
Further details on learning rates, weight decay and batch size can be found in the supplementary material. For evaluations we follow the configurations from prior works. Namely, CoTracker \cite{karaev2023cotracker} uses windows of 8 frames long with an overlap of 4 frames whilst PIPs++ \cite{zheng2023pointodyssey} uses a window for 128 frames with no overlap. All evaluations on CoTracker~\cite{karaev2023cotracker} (as well as CoTracker v3~\cite{karaev2024cotracker3}) do not use a support grid. Unless mentioned otherwise, we use a resolution of 512x384 for all our evaluations.

\noindent \textbf{Evaluation metrics.} 
We use the standard metrics previously used for for point tracking, and also introduce several new ones, to measure re-identification capabilities of point trackers. 
The metrics can broadly be separated into three categories: $\delta$ metrics, binary accuracy metrics, and error.

\begin{itemize}[noitemsep,topsep=0pt,leftmargin=*]
    \item $\delta$ accuracy metrics:
    \begin{itemize}[noitemsep,topsep=0pt,leftmargin=*]
        \item $\delta_{\text{avg}}$: average percentage of points that fall within a set of pixel thresholds. The average over multiple thresholds allows better capturing of performance improvements. We follow previous works \cite{zheng2023pointodyssey, karaev2023cotracker, doersch2022tap} and use the threshold set $\{1,2,4,8,16\}$.
        \item $\delta^*_{\text{avg}}$: Due to the low performance on EgoPoints for current models, we propose a more relaxed version, $\delta_{\text{avg}}^*$, with thresholds $\{8, 16, 24\}$.
        \item ReID$\delta_{\text{avg}}$: the percentage of correctly tracked re-identification points (see Fig~\ref{fig:re_id_metrics}). These should be correctly tracked as out-of-view then correctly tracked within a threshold from the ground truth in the final frame. We also average the threshold set $\{8,16,24\}$.
    \end{itemize}
    \item Binary accuracy metrics:
        \begin{itemize}[noitemsep,topsep=0pt,leftmargin=*]
        \item In-View Accuracy (IVA): percentage of ground-truth in-view points correctly predicted to be in-view.
        \item Out-of-View Accuracy (OOVA): percentage of ground-truth OOV points correctly predicted to be out-of-view.
        \item Occlusion Accuracy (OA): percentage of points that are correctly predicted as visible/invisible.
    \end{itemize}
    \item Median Trajectory Error (MTE): Previously used in \cite{zheng2023pointodyssey}, it measures the median of the L2 distance between the predictions and the ground truth for each track, averaged over all tracks.
\end{itemize}

\noindent \textbf{Fine-tuning results on K-EPIC.} The results in Table~\ref{tab:sota_improv} first present the performance of models on EgoPoints, across all metrics. CoTracker particularly struggles for ReID and PIPs++ performs poorly on OOVA.
We then present results of fine-tuning these models using the K-EPIC data.
Fine-tuning
yields noticeable performance improvements across all metrics for both models.  In particular, for PIPs++, more points are correctly tracked OOV, highlighted by a $1.6$ point increase in OOVA, as well as more points are successfully recovered when coming back into the frame, made clear by a $2.8$ point increase in overall ReID$\delta_{\text{avg}}$. The performance improvement for CoTracker\footnote{We use no support grid and single\_point=False} after fine-tuning on K-EPIC is even significantly larger across all metrics. The tracking accuracy show great improvement, with $2.7$ points increase in $\delta_{\text{avg}}^*$ and a reduction of $11.6$ in MTE. Although OOVA does not increase much, the number of points that return to the screen and are accurately positioned (recorded by IVA and ReID$\delta_{\text{avg}}$) show a sizeable improvement after fine-tuning. 

We presented qualitative examples of improved performance in Fig~\ref{fig:main_qual}. In the second example, points on the sink are correctly tracked after fine-tuning, with points on the draining board also correctly tracked during the sequence.

\noindent \textbf{Scene points only vs. K-EPIC.} We carry out an ablation experiment on the effect of the scene points and the dynamic points when fine-tuning models. The second row for each model in Table~\ref{tab:sota_improv} highlights the performance of fine-tuning when just using the scene points. For both PIPs++~\cite{zheng2023pointodyssey} and CoTracker~\cite{karaev2023cotracker}, nearly all metrics improve (apart from OOVA on CoTracker and OOVA and IVA on PIPs++, which actually get marginally worse), seeing improvements of $3.8$ and $0.9$ points on the ReID$\delta_{\text{avg}}$, respectively. 
As expected, in-view accuracy (IVA) is best improved as scene points are likely to be in-view during these sequences.
The best performance is achieved when using the full K-EPIC training set (with dynamic objects) for fine-tuning. 
This highlights that although the background points offer improvement from domain-specific features and movement, the dynamic objects supplement this further by introducing challenging occlusions and 3D motions.

\begin{table*}[]
\centering
\resizebox{0.7\linewidth}{!}{
\begin{tabular}{lllllllllll}
\toprule
&&\multicolumn{4}{c}{TAP-Vid-DAVIS} &&\multicolumn{4}{c}{TAP-Vid-KINETICS}\\
\cmidrule(lr){3-6} \cmidrule(lr){8-11}
Model                     && $\delta_{\text{avg}}\uparrow$ &$\delta_{\text{avg}}^*\uparrow$ & OA$\uparrow$  & AJ$\uparrow$ && $\delta_{\text{avg}}\uparrow$ &$\delta_{\text{avg}}^*\uparrow$ & OA$\uparrow$  & AJ$\uparrow$ \\
\midrule
CoTracker               & & \textbf{74.7} & 93.1& 88.7     & \textbf{60.7}    && 61.9 & 82.4& 83.1   &  \textbf{48.3}     \\
CoTracker w. Fine-Tuning && 74.3 & \textbf{93.9} & \textbf{89.0}     & 60.5    
&& \textbf{62.7}& \textbf{84.5}& \textbf{83.3}   &  48.1      \\
\bottomrule
\end{tabular}}
\caption{Results with CoTracker \cite{karaev2023cotracker} before and after fine-tuning with K-EPIC  on third person bechmarks. AJ here stands for average jaccard - a metric to measure both visibility and tracking accuracy \cite{karaev2023cotracker}.}
\vspace*{-12pt}
\label{tab:cotracker_davis}
\end{table*}

\begin{table}[]
\resizebox{\linewidth}{!}{
\begin{tabular}{lccccccc}
\toprule
&\multicolumn{3}{c}{TAP-Vid-DAVIS} &\multicolumn{3}{c}{TAP-Vid-KINETICS}\\
\cmidrule(lr){2-4} \cmidrule(lr){5-7}
Model                 & $\delta_{\text{avg}}\uparrow$ &$\delta_{\text{avg}}^*\uparrow$  & MTE$\downarrow$  & $\delta_{\text{avg}}\uparrow$ &$\delta_{\text{avg}}^*\uparrow$  & MTE$\downarrow$ \\
\midrule
PIPs++                & 64.0 & 88.8   & \textbf{7.7}  & \textbf{45.6} & 61.7  &  65.6     \\
PIPs++ w. FT & \textbf{64.6} &\textbf{89.5}   & 7.9   & 44.8 & \textbf{61.9}  &  \textbf{65.5}     \\
\bottomrule
\end{tabular}}
\vspace*{-4pt}
\caption{Results with PIPs++ \cite{zheng2023pointodyssey} before and after fine-tuning with K-EPIC on third person bechmarks.}
\vspace*{-8pt}
\label{tab:pips_davis}
\end{table}

\noindent \textbf{Maintaining performance on other datasets.} When fine-tuning there is always the danger of damaging prior performance of the models. In order to test for this, we evaluate the fine-tune models on the popular TAP-Vid-DAVIS and TAP-Vid-KINETICS evaluation datasets \cite{doersch2022tap}. As can be seen from both Tables \ref{tab:cotracker_davis} and \ref{tab:pips_davis} we manage to retain the performance of the baseline models. This, we believe, is in part due to our inclusion of the pre-training datasets during fine-tuning as well as the challenging 3D sampled points and occluded tracks of K-EPIC. 

For TAP-Vid-DAVIS, the fine-tuned CoTracker model only drops by $0.4$ points on $\delta_{\text{avg}}$ whilst increasing by $1.1$ and $2.7$ on the EgoPoints $\delta_{\text{avg}}$ and $\delta_{\text{avg}}^*$ metrics, respectively. A similarly small drop is seen in the average jaccard (AJ) metric whilst a small increase is actually seen in the visibility accuracy. For the PIPs++ fine-tuned model on the other hand, both $\delta_{\text{avg}}$ and $\delta_{\text{avg}}^*$ metrics increase whilst the MTE increases by a small margin of $0.2$. For TAP-Vid-KINETICS, we similarly retain performance across most of the metrics for both fine-tuned models. For both datasets and models, fine-tuning on K-EPIC clearly maintains performance on while consistently improving performance on egocentric videos, as shown in Table \ref{tab:sota_improv}.

\begin{table}[]
\resizebox{\linewidth}{!}{
\begin{tabular}{lcccc}
\toprule
               & \multicolumn{2}{l}{Scene} & \multicolumn{2}{l}{Objects} \\
\midrule
Model        & $\delta_{\text{avg}}\uparrow$    & ReID$\delta_{\text{avg}}\uparrow$     & $\delta_{\text{avg}}\uparrow$     & ReID$\delta_{\text{avg}}\uparrow$     \\
\midrule
PIPs++      & 69.2 & 19.1 & \textbf{38.2} & 11.1 \\
PIPs++ w FT & \textbf{69.7} & \textbf{19.9} & 37.9 & \textbf{12.3} \\
\midrule
CoTracker      & 56.7 & 5.1 & 35.9 & 1.6 \\
CoTracker w FT & \textbf{61.0} & \textbf{8.4} & \textbf{41.2} & \textbf{5.4} \\
\bottomrule
\end{tabular}}
\caption{Ablation on EgoPoints, reporting results separately for scene points and dynamic object points.}
\vspace*{-12pt}
\label{tab:dyn_stat_split}
\end{table}

\noindent \textbf{Scene points vs dynamic object points.} As a further ablation of the fine-tuning, we report results separately for the scene and dynamic objects points. These have been flagged by our annotator for 75.8\% of all tracks annotated.
In total, the results from Table \ref{tab:dyn_stat_split} cover 2149 scene tracks and 1414 dynamic object tracks. As can be seen from these results, there is a clear improvement for both tracking accuracy and re-identification on the scene points for both PIPs++ and CoTracker when using our fine-tuning strategy. Furthermore, we see an improvement for re-identification on the dynamic tracks. A drop in performance on object points while fine-tuning PIPs++ could be explained by the original PIPs++ model's tendency to keep points in-view.
Of particular note is CoTracker's failure to ReID nearly all object points before being fine-tuned ({ReID$\delta_{\text{avg}}$ = 1.6}), improving to 5.4 after fine-tuning.

\begin{figure}[t]
  \centering
   \vspace*{-8pt} 
\includegraphics[width=\linewidth]{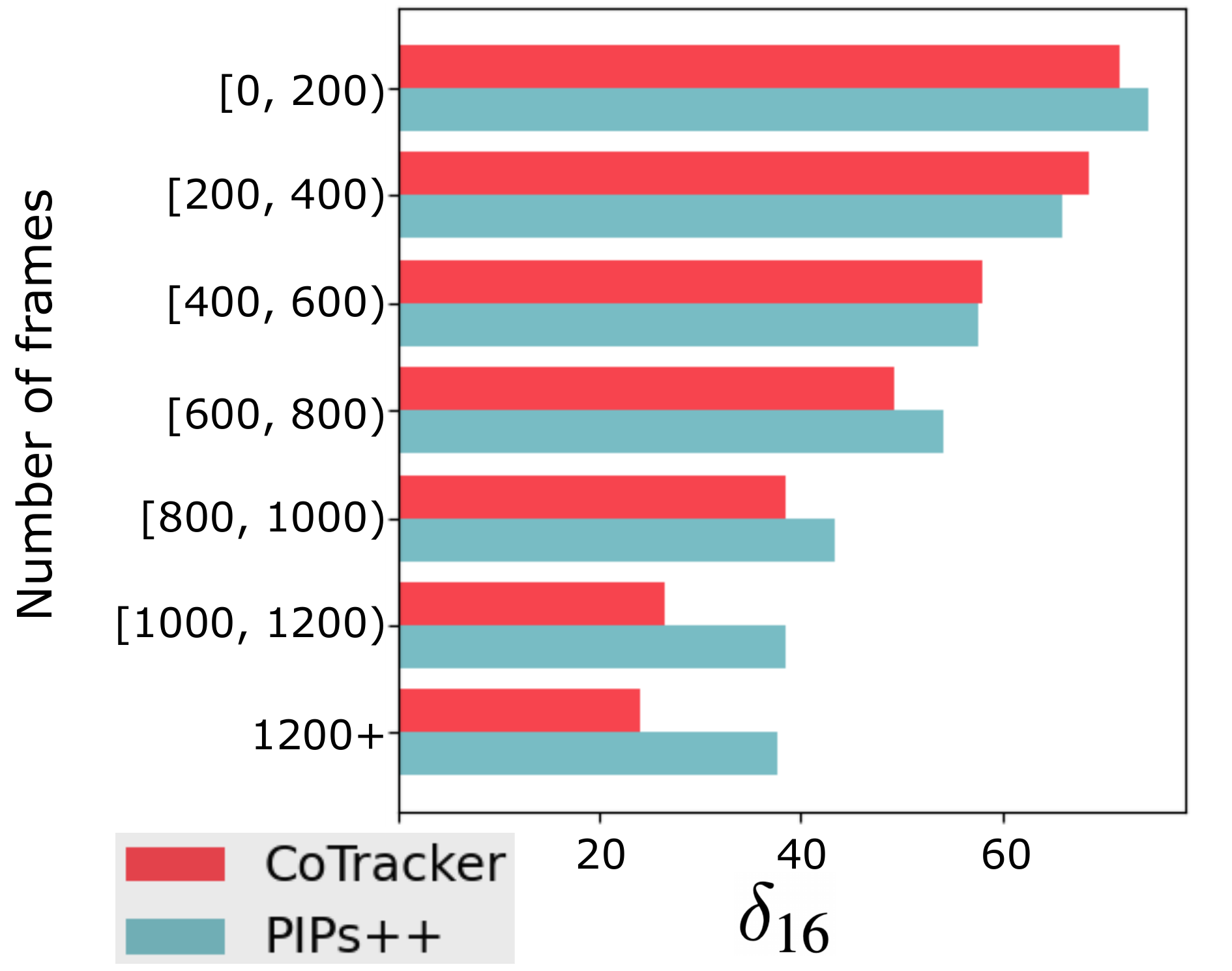}

   \caption{Average $\delta_{16}$ across varying sequence lengths on PIPs++~\cite{zheng2023pointodyssey} and CoTracker \cite{karaev2023cotracker}, after fine-tuning on K-EPIC.}
   \vspace*{-16pt}
   \label{fig:d_x_by_len}
\end{figure}

\noindent \textbf{Performance vs sequence length.}  Figure \ref{fig:d_x_by_len} shows $\delta_{16}$ for fine-tuned models of PIPs++ \cite{zheng2023pointodyssey} and CoTracker \cite{karaev2023cotracker}. 
Both models perform best at shorter videos (0-200 frames long), dropping performance steadily as sequence length increases.
Interestingly, CoTracker struggles more for the longer sequences ($>$1K frames).

\noindent \textbf{Limitations.} Through introducing the re-identification and view metrics (ReID, IVA, OOVA), we showcase current point tracking methods' limitations in tracking both scene and dynamic objects leaving the field of view and re-appearing.
We demonstrate that synthetic sequences could be designed to improve ReID performance, however performance remains modest with most point tracking failing to ReID points.
The best overall performance on ReID$\delta_{\text{avg}}$ reported in this paper is 16.8\% (Table~\ref{tab:sota_improv}). This performance, while improving prior work, showcases ReID as a major obstacle to point tracking, particularly in egocentric videos.
Algorithmic improvements are required to further address the shortcomings of SOTA methods.

\section{Conclusion}
\label{sec:conclusion}
In this paper, we introduce, for the first time, point tracking for egocentric videos. The paper offers contributions in three areas. 
First, we introduce EgoPoints, a benchmark of annotated 4.7K challenging point tracks in 517 egocentric sequences. The benchmark includes flags that allow analysing points that are on scene objects, dynamic objects, those that are in-/out-of-view and points that need to be re-identified on return.
Second, we analyse performance of SOTA point tracking models on egocentric videos. We introduce metrics to particularly analyse the ability to re-identify points, which is a frequent challenge in egocentric videos.
Third, we propose a pipeline to create semi-real sequences with automatic annotations for fine-tuning models. These sequences combine scene points, from camera estimates of egocentric sequences, with 3D object points, from synthetic 3D models. Fine-tuning improves performance on egocentric videos,
while maintaining performance on popular third-person point tracking benchmarks. 

\small{\noindent \textbf{Acknowledgments:} This work proposes a new annotations benchmark that is publicly available, and builds on publicly available dataset EPIC-KITCHENS. It is supported by EPSRC Doctoral Training Program, EPSRC UMPIRE EP/T004991/1 and EPSRC Programme Grant VisualAI EP/T028572/1. We acknowledge the use of the EPSRC funded Tier 2 facility \mbox{JADE-II}.}

{\small
\bibliographystyle{ieee_fullname}
\bibliography{refs}
}

\newpage 
\renewcommand{\thesubsection}{\Alph{subsection}}
\section*{Appendix}
\subsection{Implementation Details}

Here we provide the required additional implementation details to replicate our results.

\noindent \textbf{PIPs++.} We adopt the $200k$ iterations checkpoint provided by~\cite{zheng2023pointodyssey}, which is pre-trained on the PointOdyssey dataset. We fine-tune for a further $45k$ iterations using the data mix described in the main paper. Specifically, for each batch, there is a 65\% chance of sampling K-EPIC sequences (where half of these are looped for increased re-identifications) and a 35\% chance of sampling from the original PointOdyssey training dataset. We use a sequence length of 36 frames for PointOdyssey and 24 frames for K-EPIC. We resize K-EPIC sequences to 384x512. We use a batch size of 2, 128 trajectories per sequence and a constant learning rate of $2.8e^{-7}$ on a single V100 32GB GPU.

\noindent \textbf{CoTracker~ \cite{karaev2023cotracker}.} We make use of the CoTracker-v2 checkpoint provided by the authors. This was trained for $50k$ iterations on sequences of 24 frames from the TAP-Vid-KUBRIC dataset \cite{doersch2022tap} and utilises the virtual tracks added in the second version of the work. We then fine-tune this model further with the same data mix as PIPs++ above, between K-EPIC and TAP-Vid-KUBRIC. We use a batch size of 1, 196 trajectories per sequence and a learning rate of $5e^{-5}$ with a linear 1-cycle\footnote{Leslie N Smith and Nicholay Topin. Super-convergence: Very fast training of neural networks using large learning rates. In Artificial Intelligence and Machine Learning for Multi-Domain Operations Applications, volume 11006, page 1100612. International Society for Optics and Photonics, 2019} learning rate schedule following CoTracker training. We use two V100 32GB GPUs. We train CoTracker with virtual tracks of 64 following the provided code~\cite{karaev2023cotracker}.

\noindent \textbf{CoTracker3 \cite{karaev2024cotracker3}.}
Similar to CoTracker-v2, we evaluate CoTracker3 at 384x512 resolution. We use the pre-trained online model provided by the authors. 

\noindent \textbf{LocoTrack \cite{cho2024local}.}
We evaluate models on their native (training) resolution which is 256x256. Due to memory constraints, we set a maximum limit of 1,000 frames during inference. For sequences exceeding this limit, we sample equally spaced frames ensuring we always include the annotated frames.

\noindent \textbf{BootsTAPIR Online \cite{doersch2024bootstap}.}
We use the sequential, causal version of BootsTAPIR, implemented in PyTorch and provided at the official github\footnote{https://github.com/google-deepmind/tapnet}
\begin{figure*}[!h]
  \centering
    \includegraphics[width=1.0\linewidth]{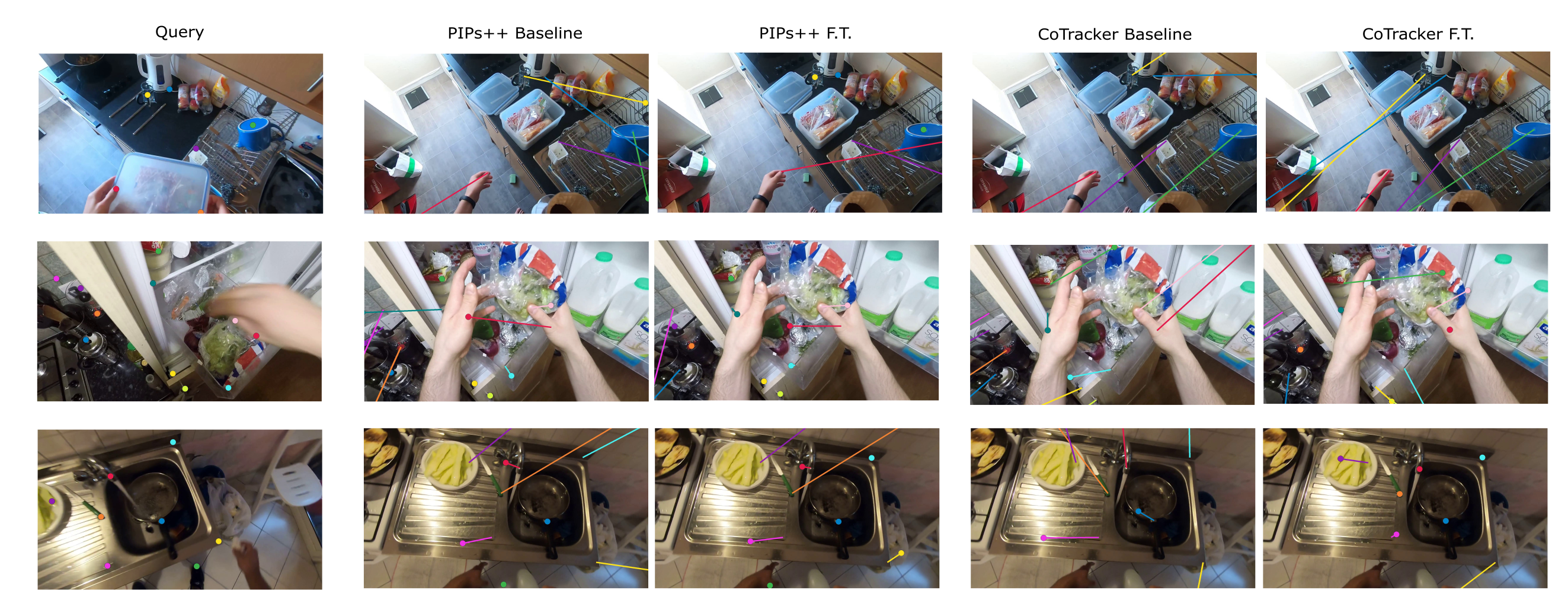}
    \caption{Three examples of EgoPoints evaluations before and after fine-tuning on PIPs++ and CoTracker. Dots represent initial points in the first column, and predictions in the other four columns. We plot a line connecting the prediction to the ground truth so as to show the difference. Points are correctly predicted if no line is attached. Points connected to the image boundary indicates the point is predicted out-of-view.}
  \label{fig:annot_comp}
\end{figure*}

\subsection{Qualitative Examples}

Three examples of predictions on EgoPoints annotations for both PIPs++ \cite{zheng2023pointodyssey} and CoTracker \cite{karaev2023cotracker}, before and after fine-tuning, can be seen in Figure \ref{fig:annot_comp}. It should be noted that we show the first and final evaluation frames for simplicity. However, each of these examples involve the camera wearer moving around the scene before revisiting the same location in the first frame. Therefore, they are particularly difficult re-identification scenarios for current SOTA models, as discussed in the main paper.

These examples demonstrate a clear improvement over the baselines. The first two examples are good examples of where PIPs++ \cite{zheng2023pointodyssey} does better at re-identification than CoTracker \cite{karaev2023cotracker}. The first example is 830 frames long and it is possible to see that fine-tuning on PIPs++ helps to successfully re-identify the yellow, blue and green points that were lost by the baseline.

The second example is another case of where PIPs++ improves more than CoTracker when fine-tuning. The dark purple, orange and dark green points are all successfully recovered when compared to the baseline. For CoTracker, although the orange and dark green points are tracked correctly after fine-tuning, while points at the bottom of the frame are lost.

The third sequence shows CoTracker performing better after fine-tuning. 5 of the 8 query points are tracked precisely and a sixth point (the dark purple) is tracked close to the ground truth.

\begin{figure*}[!h]
  \centering
    \includegraphics[width=1.0\linewidth]{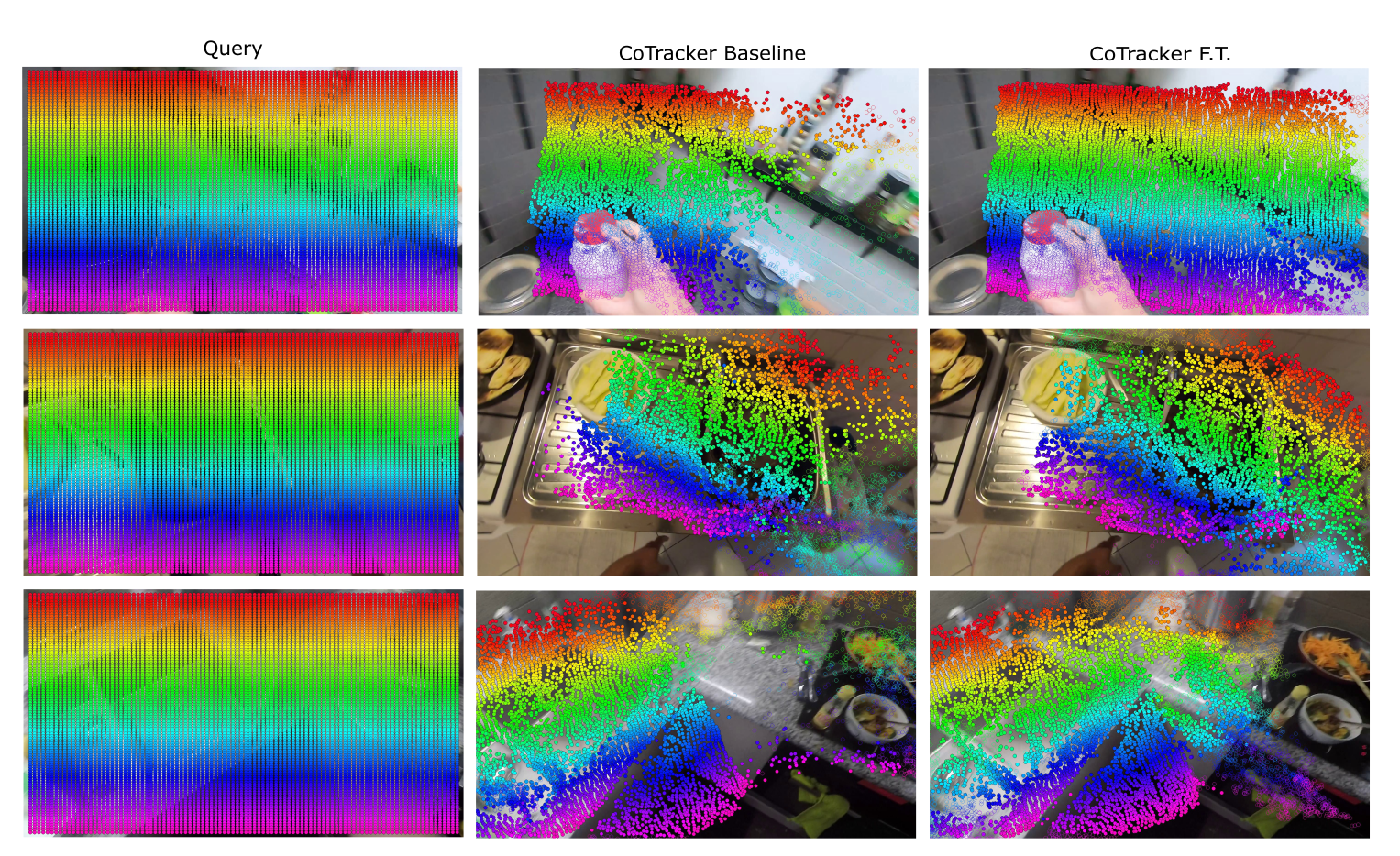}
    \caption{Examples of CoTracker results, before and after fine-tuning, on a dense grid of points (100x100).}
  \label{fig:grid_comp}
\end{figure*}

We also show qualitative results using dense query grids for 
CoTracker~\cite{karaev2023cotracker} in Figure \ref{fig:grid_comp}. In all examples the baseline can be seen to struggle with the complete grid during re-identification. After fine-tuning with K-EPIC, most points are recovered. 
We share a video of these sequences on the project webpage.

\begin{figure*}[!h]
  \centering
    \includegraphics[width=1.0\linewidth]{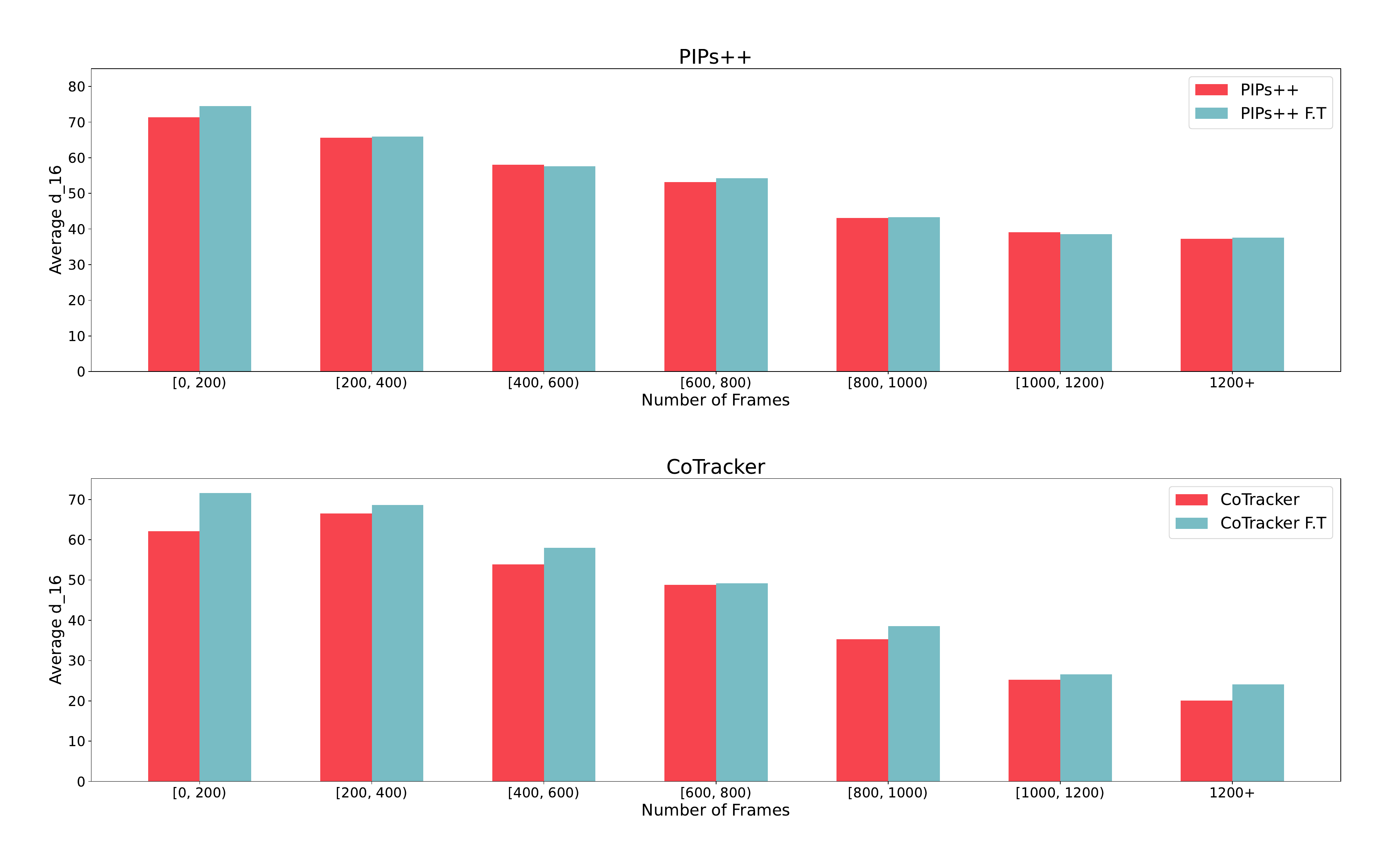}
    \vspace*{-24pt}
    \caption{Average $\delta_{16}$ vs. sequence length of EgoPoints benchmark, before and after fine-tuning on PIPs++ \cite{zheng2023pointodyssey} and CoTracker \cite{karaev2023cotracker}.}
  \label{fig:figure_d_x_by_len_v5}
\end{figure*}
In the main paper, we ablate the performance of fine-tuned models over sequence lengths.
As an extension to this, we show here that fine-tuning improves performance across sequence lengths. Figure \ref{fig:figure_d_x_by_len_v5} shows average $\delta_{16}$ for ranges of 200 frames.
For PIPs++, performance is improved for short sequences ($<$ 200 frames) with comparable performance for sequences (2K-2.2K frames in length). 
On the other hand, CoTracker shows clearer improvements throughout.

\end{document}